\documentclass[10pt]{IEEEtran}
\usepackage[utf8]{inputenc}
\usepackage{fancyhdr}
\usepackage{lastpage}
\usepackage{a4wide} 
\usepackage{amsmath,amsthm}
\usepackage{amssymb} 
\usepackage{graphicx}
\usepackage{color}
\usepackage{fancybox}
\usepackage{verbatim}
\usepackage{moreverb}
\usepackage{listings}
\usepackage{hyperref}
\usepackage{inputenc}
\usepackage[ruled,vlined]{algorithm2e}
\usepackage{pgf,tikz}
\usetikzlibrary{babel}

\newtheorem{remark}{Remark}

\usepackage{geometry}
\usepackage{pgfplots}
\usepgfplotslibrary{fillbetween}
\pgfplotsset{width=10cm,compat=1.9}

\title{Multi-task learning on the edge:\\ cost-efficiency and theoretical optimality}
\author{Sami Fakhry$^1$, Romain Couillet$^{1,2,\star}$, Malik Tiomoko$^1$ \\ $^1$GIPSA-lab, $^2$LIG-lab, Grenoble-Alps University, France.\thanks{\noindent$^\star$Couillet's work is supported by the MIAI LargeDATA chair at Univ.\@ Grenoble-Alps.} }
\date{July 2021}

\begin{document}

\maketitle

\section*{\centering {Abstract}}

This article proposes a distributed multi-task learning (MTL) algorithm based on supervised principal component analysis (SPCA) \cite{bair2006prediction,barshan2011supervised}, which is: (i) theoretically optimal for Gaussian mixtures, (ii) computationally cheap and scalable. Supporting experiments on synthetic and real benchmark data demonstrate that significant energy gains can be obtained with no performance loss.

\section{Introduction}

The mandatory low carbon-footprint revolution in technologies impacts our ``consumption'' of data storage, exchange, and computational power. In this view, transfer and multi-task learning \cite{caruana1997multitask,zhang2018overview,torrey2010transfer,yang2020transfer} are efficient solutions to exploit remotely located datasets, but their optimal designs in general demand to gather all data together. In parallel, edge computing \cite{khan2019edge,shi2016promise,satyanarayanan2017emergence,varghese2016challenges} proposes to maintain computations locally with minimal exchanges but at the expense of performance.

We introduce here a cost-efficient ``multi-task learning on the edge'' paradigm which draws the strengths of minimalistic data exchanges from edge computing and of multi-task learning to exploit multiple (possibly statistically distinct) datasets together. By exchanging the \emph{sufficient statistics} (rather than the data) and by optimizing the task-dependent data labels (rather than setting them all to $\pm 1$ as conventionally done), the proposed scheme is proved information-theoretically optimal.

\medskip

Specifically, the article provides a distributed and scalable extension of the supervised PCA-based multi-task learning algorithm (MTL-SPCA) \cite{MTLSPCA}. For large and numerous data, the resulting algorithm is provably equivalent in performance to the original (and Gaussian-optimal) MTL-SPCA, while simultaneously allowing for drastic cost reductions in data sharing. 
In particular, the algorithm recovers all advantages from MTL-SPCA, such as the absence of the deleterious problem of \emph{negative transfers}. 

\smallskip

\noindent{\bf Reproducibility.} Code of all figures and algorithms provided in the article are available at {\color{blue}\url{https://github.com/Sami-fak/DistributedMTLSPCA}}.


%

\section{Centralized Algorithm}
\label{sec:centralized}

MTL-SPCA \cite{MTLSPCA} is a fast supervised multi-task classification algorithm relying on a preliminary PCA-like step which projects the data onto a \emph{subspace of weighted data classes}. Similar to PCA, MTL-SPCA is a mathematically tractable spectral method. 
This tractability allows for setting \emph{theoretically optimal data labels} (rather than $\pm1$), these \emph{labels not being intrinsic to the data but differing for each target task} so to minimize the classification error for each individual task. Of utmost importance here, the inner functioning of MTL-SPCA relies on ``condensed'' data information arising from each task (the class-wise statistical means of the data), thereby easily allowing for a distributed extension.

Consider $k$ independent agents (i.e., \emph{clients}) solving a classification task on their own data based on training inputs $X_{1},\ldots,X_{k}$, where $X_i\in\mathbb R^{p\times n_i}$ is the dataset of client (or \emph{task}) $i$ composed of $n_i$ data of size $p$. 
We let $X=[X_{1},\ldots,X_{k}]\in\mathbb{R}^{p\times n}$ ($n=\sum_{t=1}^kn_t$) and, for task $i$, $X_i=[X_{i1},\ldots,X_{im}]$ for $X_{ij}\in\mathbb R^{p\times n_{ij}}$ the class-$j$ subset ($1\leq j\leq m$) for client $i$ (so $n_i=\sum_{j=1}^m n_{ij}$). With a one-versus-all approach, the $m$-class problem can be decomposed as a series of $m$ $2$-class problems: for readability, we thus restrict ourselves to $m=2$ classes.

We further denote, for class $j$ in task $i$, $X_{ij}=[x_{i1}^{(j)},\ldots,x_{in_{ij}}^{(j)}]\in\mathbb{R}^{p\times n_{ij}}$. To each $x_{i\ell}^{(j)}$ is classically associated a label $y_{i\ell}^{(j)}\in\{\pm1\}$. In \cite{MTLSPCA}, this choice is proved largely suboptimal and the main source of negative transfers. Instead, for large $p,n_{ij}$, if all data in $X$ are independent Gaussian vectors with covariance $I_p$, the label $y_{i\ell}^{(j)}$ minimizing the classification error \emph{for target task $t$} should be taken constant for all $\ell$ and equal to $y_{i\ell}^{(j)}=[\tilde y^{[t]}]_{ij}$ where $\tilde y^{[t]}\in\mathbb R^{2k}$ is the vector: $$\tilde{y}^{[t]}=\mathcal{D}_c^{-\frac12}(\mathcal{M}+I_{2k})^{-1}\mathcal{M}\mathcal{D}_c^{-\frac12}(e_{t1}-e_{t2})\in\mathbb{R}^{2k},$$ 
here $e_{tj}=e_{2(t-1)+j}\in\mathbb{R}^{2k}$ is the canonical vector with 1 at the $2(t-1)+j$-th coordinate,
$$\mathcal{M}=(1/c_0)\mathcal{D}_c^{\frac12}M^{\sf T}M\mathcal{D}_c^{\frac12}\in\mathbb{R}^{2k\times 2k}$$ 
is the (fundamental) inter-task similarity matrix, 
$$c=[n_{11}/n,\ldots, n_{k2}/n]^T\in\mathbb{R}^{2k}$$
the vector of size ratios with $\mathcal{D}_c={\rm diag}(c)$, $c_0=p/n$, and $M$ the matrix of statistical means
$$
M = [\mu_{11},\mu_{12},\ldots,\mu_{k1},\mu_{k2}],\quad \mu_{tj}=\mathbb E[x_{t1}^{(j)}].
$$
Note importantly that the optimal label vector $\tilde y^{[t]}$ for task $t$ \emph{depends on the target task}.

\begin{remark}[Estimating $\tilde y^{[t]}$]
\label{rem:estimate_y}
For large Gaussian data, $M^{\sf T}M$ can be effectively estimated as \cite{MTLSPCA}
\begin{align*}
    [M^{\sf T}M]_{qq} &= \frac4{n_{ij}^2}\mathbf{1}_{n_{ij}}^{\sf T}X_{ij;1}^{\sf T}X_{ij;2}\mathbf{1}_{n_{ij}} + o_p(1) \\
    [M^{\sf T}M]_{qq'} &= \frac1{n_{ij}n_{i'j'}}\mathbf{1}_{n_{ij}}^{\sf T}X_{ij}^{\sf T}X_{i'j'}\mathbf{1}_{n_{i'j'}} + o_p(1)
\end{align*}
with $q=2(i-1)+j$ and $q'=2(i'-1)+j'$ different and $X_{ij}=[X_{ij;1},X_{ij;2}]$ an even-sized division of $X_{ij}$. With this result, $\hat{\tilde y}^{[t]}$, defined as $\tilde y^{[t]}$ with $M^{\sf T}M$ replaced by the above estimates, is a consistent estimate for $\tilde y^{[t]}$ as $p,n_{ij}$ grow large.
\end{remark}

Remark~\ref{rem:estimate_y} shows that the optimal $\tilde y^{[t]}$ is empirically accessible from the vectors $\frac1{n_{ij}}X_{ij}\mathbf{1}_{n_{ij}}$: it does not require to know the individual data.


\medskip


To classify a new sample $\bf x$ for client $t$, MTL-SPCA then consists in projecting $\bf x$ onto the vector
$$ V_t = {XJ \tilde{y}^{[t]}}/{\|XJ \tilde{y}^{[t]}\|} = {Xy}/{\|Xy\|} $$
with $J=\big[j_{11}, \ldots, j_{2k} \big]$, where $j_{tj} = \big[0,\ldots,0,\mathbf{1}_{n_{tj}},0,\ldots,0 \big]^T$, and $y=J\tilde{y}^{[t]}$.
Precisely, the error-rate minimizing decision (in a Gaussian setting) results from the test
\begin{align}
\label{eq:gt}
    g_t({\bf x}) \equiv V_t^{\sf T}{\bf x} - \zeta_t \gtrless 0,~ \zeta_t \equiv \frac12(\hat{m}_{t1}+\hat{m}_{t2})
\end{align}


Here again, $Xy=\sum_{ij}[\tilde{y}^{[t]}]_{ij}X_{ij}{\bf 1}_{n_{ij}}$ is a linear combination of the empirical means $\frac1{n_{ij}}X_{ij}{\bf 1}_{n_{ij}}$.



\section{Distributed Algorithm}
\label{sec:distributed}

All operations leading to $g_t(\cdot)$ can be written as a function of the empirical means
$$\hat{\mu}_{ij}\equiv \frac1{n_{ij}}X_{ij}{\bf 1}_{n_{ij}},\quad i\in\{1,\ldots,k\},~j\in\{1,2\}.$$
The complete dataset $X$, thus, \emph{needs not to be accessible to client $t$}: only the $\hat\mu_{ij}$'s are necessary and MTL-SPCA can be fully distributed by only sharing the local (estimated) statistical means across clients. Or, more efficiently, by updating a set of statistical means within a \emph{central client}, to and from which every client may upload the $\hat\mu_{ij}$'s and download $V_t$. To get rid of the dataset dependence tied to the centralized version, $V_t = {Xy}/{\|Xy\|}$ is here rewritten using:
\begin{align}Xy 
&=\sum_{i=1}^k\sum_{j=1}^2n_{ij}\hat{\mu}_{ij} [\tilde y^{[t]}]_{ij}.
\end{align}

\medskip

\begin{figure}[h!]
    \centering
    \includegraphics[scale=.2]{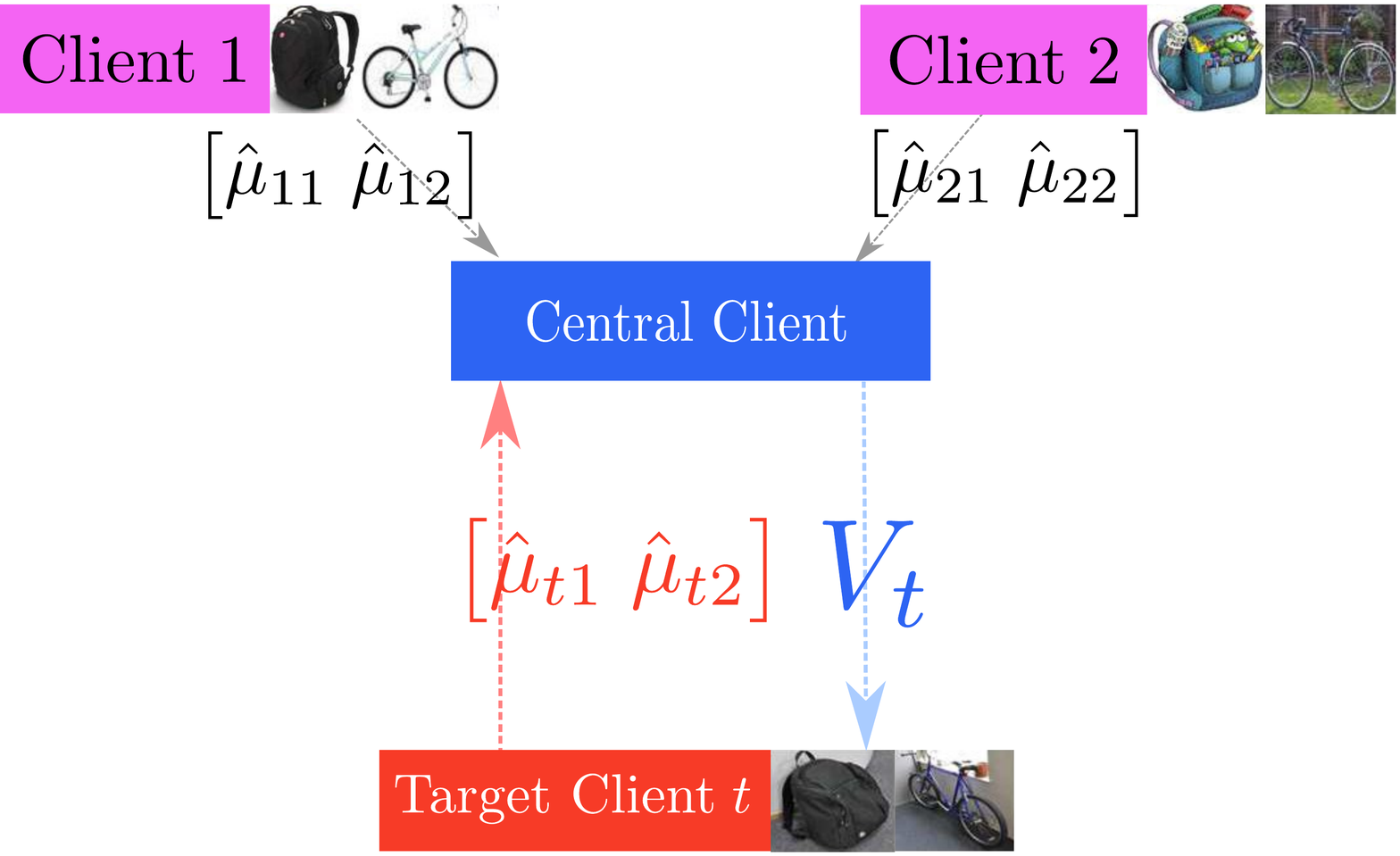}
        \vspace*{-.5cm}
    \caption{Schematics of the distributed algorithm for the Amazon (Client 1), Caltech (Client 2), and Webcam (Target Client $t$) database. Client $t$ draws the projection vector $V_t$ from the empirical statistical means shared by all clients.}
    \label{fig:summary}
\end{figure}

The distributed approach is depicted in Figure~\ref{fig:summary} for the popular multitask learning Amazon-Caltech-DSLR-Webcam database. Every client is a data center. Target client ``webcam'' aims to classify images of bags and bikes, exploiting images of the same objects from other data centers slightly differing in their features (number of data, size, shape, resolution, background, diversity, etc). Distributed MTL-SPCA only requires here for the ``webcam'' (target) client to access the empirical means from the other two data centers. 


\medskip
Explicitly, using Remark~\ref{rem:estimate_y} applied to all constituents of $g_t(\cdot)$ -- that is, writing $Xy$ and the $\hat{m}_{ij}$'s 
as functions of the $\hat{\mu}_{ij}$'s --, MTL-SPCA is distributed as per Algorithm~\ref{alg:client} (by the central client) and Algorithm~\ref{alg:target} (by target task $t$).
\begin{algorithm}[h!]
\SetAlgoLined
\KwData{$\hat M_1,\ldots\hat M_k$ ($\hat M_i=[\hat{\mu}_{i1},\hat{\mu}_{i2}]$), for $\hat\mu_{ij}\equiv\frac{1}{n_{ij}}X_{ij}\mathbf{1}_{n_{ij}}$, gathered as $\hat{M}=[\hat M_1,\ldots,\hat M_k]\in\mathbb R^{p\times 2k}$\; 
}

 \textbf{Estimate} similarity $\mathcal{M}$, optimal label $\tilde{y}^{[t]}$, and then projection $V_t$ and decision threshold $\zeta_t$ (Remark~\ref{rem:estimate_y})\;
 \textbf{Send} estimates of $V_t,\zeta_t$ to client $t$.
 \caption{Central Client}
 \label{alg:client}
\end{algorithm}
\vspace*{-.6cm}
\begin{algorithm}[h!]
\SetAlgoLined
\KwData{Training $X_{t}$ for task $t$. Test data $\mathbf{x}$.}
\KwResult{Class $\hat{j}\in\{1,2\}$ of $\mathbf{x}$ for task $t$.}
 \textbf{Compute} class empirical means $\hat{\mu}_{t1},\hat{\mu}_{t2}$\; 
 \textbf{Send} $M_t=[\hat{\mu}_{t1},\hat{\mu}_{t2}]$ to \emph{Central Client}\;
 \textbf{Retrieve} $V_t,\zeta_t$ from \emph{Central Client}\;
 \textbf{Compute} score $g_t({\bf x})$ and classify $\mathbf{x}$ in class $\hat j$ from decision $g_t({\bf x})\underset{\hat j=2}{\overset{\hat j=1}{\gtrless}} \zeta_t$.
 \caption{Target task $t$}
 \label{alg:target}
\end{algorithm}





\begin{remark}[Transmission costs]
A decisive advantage of MTL-SPCA is to allow for a ``multi-task on the edge'' approach, drastically reducing computational costs compared to a centralized implementation. In operation, the $k$ clients only send $(m-1)k$ size-$p$ vectors ($m-1$ per client) to a \emph{Central Client},
thereby preventing the transfer of the complete datasets (which would require $\sum_{t,c}n_{tc}=n$ size-$p$ vectors): so a $O(n)$-fold gain. This ``compacity'' of the sufficient statistics also ensures data privacy by emitting an averaged version of individual data.
\end{remark}

\section{Experiments}
\label{sec:results}

\subsection{2-class 2-task transfer learning}

In this first setting, class data would classically be labelled as $\pm1$: of course, MTL-SPCA will overwrite those by optimal labels. 
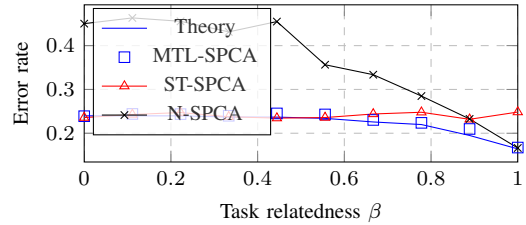
\begin{figure}[h!]
\centering
\begin{tikzpicture}
\begin{axis}[
    width=\linewidth,height=.5\linewidth,font=\footnotesize,
    xlabel={Task relatedness $\beta$},
    ylabel={Error rate},
    legend style={anchor=north west,at={(.02,.98)},font=\footnotesize,fill opacity=0.8, draw opacity=1},
    ymajorgrids=true,
    xmajorgrids=true,
    grid style=dashed,
    xmin=0,xmax=1
]

\addplot[
    color=blue
    ]
    coordinates {
    (0.0, 0.23974021347741697)(0.1111111111111111, 0.23964187860292774)(0.2222222222222222, 0.23972836899779731)(0.3333333333333333, 0.23702807823439187)(0.4444444444444444, 0.23683801736334115)(0.5555555555555556, 0.23395668307251516)(0.6666666666666666, 0.2257655662664667)(0.7777777777777777, 0.21985341567648758)(0.8888888888888888, 0.19492995441940986)(1.0, 0.16428249171959852)
    };
\addlegendentry{Theory}

\addplot [
    only marks,
    mark=square,
    color = blue,
    error bars/.cd,
    error bar style={color=blue},
    ]
    coordinates {
      (0.0, 0.23891) +- (0.0,  0.01700096761952096)(0.1111111111111111, 0.24404999999999996) +- (0.1111111111111111,  0.009069977949256534)(0.2222222222222222, 0.24367999999999998) +- (0.2222222222222222,  0.01606871494550826)(0.3333333333333333, 0.23991) +- (0.3333333333333333,  0.015227176363331457)(0.4444444444444444, 0.24525000000000002) +- (0.4444444444444444,  0.014952742223418417)(0.5555555555555556, 0.24291999999999997) +- (0.5555555555555556,  0.014527339742705819)(0.6666666666666666, 0.23034000000000004) +- (0.6666666666666666,  0.010703195784437456)(0.7777777777777777, 0.22372999999999998) +- (0.7777777777777777,  0.00979531010228875)(0.8888888888888888, 0.20973000000000003) +- (0.8888888888888888,  0.004678044463234592)(1.0, 0.16733999999999996) +- (1.0,  0.005228996079554862)
    };
    \addlegendentry{MTL-SPCA}
    
\addplot [
    mark=triangle,
    color=red,
    ]
    coordinates {
      (0.0, 0.23575)(0.1111111111111111, 0.24280999999999997)(0.2222222222222222, 0.24721)(0.3333333333333333, 0.23757)(0.4444444444444444, 0.23453)(0.5555555555555556, 0.23584)(0.6666666666666666, 0.24398)(0.7777777777777777, 0.24791000000000002)(0.8888888888888888, 0.23184)(1.0, 0.24835999999999997)
    };
    \addlegendentry{ST-SPCA}
    
\addplot [
    mark=x,
    color=black,
    ]
    coordinates {
      (0.0, 0.45016999999999996)(0.1111111111111111, 0.46325000000000005)(0.2222222222222222, 0.45550999999999997)(0.3333333333333333, 0.43142)(0.4444444444444444, 0.45489999999999997)(0.5555555555555556, 0.35639)(0.6666666666666666, 0.33365)(0.7777777777777777, 0.28493)(0.8888888888888888, 0.23260999999999998)(1.0, 0.16636)
    };
    \addlegendentry{N-SPCA}
    
\end{axis}
\end{tikzpicture}
    \vspace*{-.2cm}
\caption{Classification error for various task similarities ($\beta$) in a distributed scenario for classical label $\pm 1$ (N-SPCA), proposed method (MTL-SPCA), and single-task SPCA (ST-SPCA). The theoretical performance for the proposed method is shown in a solid line. Averaged over $10\,000$ test samples. {\bf MTL-SPCA strongly benefits from increases in task relatedness.}}
\label{fig:transfer}
\end{figure}
Figure~\ref{fig:transfer} illustrates the empirical versus theoretical (from \cite{MTLSPCA}) classification error performance of MTL-SPCA under this scenario for $x_{tl}^{(j)}\sim\mathcal{N}((-1)^j\mu_t,I_p)$, where $\mu_2=\beta\mu_1+\sqrt{1-\beta^2}\mu_1^\perp$ for any $\mu_1^\perp$ orthogonal to $\mu_1$, both of norm $\|\mu_1\|=\|\mu_1^\perp\|= 1$ with $p=100$, $n_{11}=n_{12}=1000$ and $n_{21}=n_{22}=50$. Here $\beta\in[0,1]$ enforces task relatedness. Unlike ST-SPCA (Single Task), MTL-SPCA benefits from increases in task relatedness and avoids negative transfer (N-SPCA) where labels are set to $\pm 1$.

\subsection{Adding tasks}
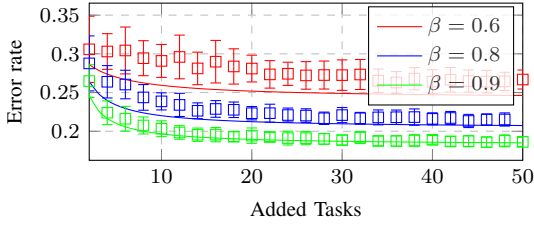
\begin{figure}[t!]
\centering
\begin{tikzpicture}
\begin{axis}[
    width=\linewidth,height=.5\linewidth,font=\footnotesize,
    legend style={anchor=north east,at={(.98,.98)},font=\footnotesize,fill opacity=0.8, draw opacity=1},    
    xlabel={Added Tasks},
    ylabel={Error rate},
    ymajorgrids=true,
    xmajorgrids=true,
    grid style=dashed,
    xmin=2, xmax=50
]

\addplot[
    color=red,smooth
    ]
    coordinates {
    (2, 0.2859844753528925)(3, 0.27888913014156413)(4, 0.2737683181429582)(5, 0.2698958976418123)(6, 0.2668638497850333)(7, 0.26442486626192774)(8, 0.2624201723460729)(9, 0.260743104747849)(10, 0.2593193269252373)(11, 0.25809543207923896)(12, 0.25703205484891256)(13, 0.25609953768860644)(14, 0.2552751100730558)(15, 0.2545409978673997)(16, 0.2538831234503604)(17, 0.25329019178485407)(18, 0.252753034986883)(19, 0.25226413389685365)(20, 0.2518172632668888)(21, 0.251407224826539)(22, 0.2510296438326115)(23, 0.2506808121549804)(24, 0.2503575659326598)(25, 0.25005718922698905)(26, 0.2497773374460147)(27, 0.24951597596216746)(28, 0.24927133051821582)(29, 0.2490418468616979)(30, 0.24882615766430716)(31, 0.24862305523689443)(32, 0.2484314688889211)(33, 0.24825044603531693)(34, 0.2480791363464495)(35, 0.24791677838422854)(36, 0.24776268828092352)(37, 0.2476162501054373)(38, 0.24747690763067087)(39, 0.2473441572698486)(40, 0.24721754199261475)(41, 0.24709664606593001)(42, 0.2469810904921667)(43, 0.24687052903890494)(44, 0.24676464477277432)(45, 0.24666314702422826)(46, 0.24656576872199953)(47, 0.24647226404576622)(48, 0.24638240635355213)(49, 0.24629598634707567)(50, 0.2462128104437571)(51, 0.2461326993287547)(52, 0.24605548666417904)(53, 0.2459810179359715)(54, 0.24590914942160308)(55, 0.24583974726405733)(56, 0.24577268663958457)(57, 0.24570785100829567)(58, 0.24564513143815525)(59, 0.24558442599410846)(60, 0.24552563918515002)(61, 0.24546868146299589)(62, 0.24541346876686104)(63, 0.24535992210943935)(64, 0.2453079671998224)(65, 0.24525753409954748)(66, 0.24520855690844584)(67, 0.24516097347731342)(68, 0.2451147251447639)(69, 0.24506975649595353)(70, 0.24502601514101802)(71, 0.2449834515114528)(72, 0.24494201867270265)(73, 0.24490167215149894)(74, 0.2448623697766032)(75, 0.2448240715317505)(76, 0.24478673941974138)(77, 0.24475033733668244)(78, 0.24471483095550023)(79, 0.24468018761796906)(80, 0.2446463762345158)(81, 0.24461336719113425)(82, 0.2445811322628889)(83, 0.2445496445333984)(84, 0.2445188783198764)(85, 0.2444888091032565)(86, 0.24445941346301947)(87, 0.24443066901634775)(88, 0.24440255436128)(89, 0.2443750490235722)(90, 0.24434813340695305)(91, 0.24432178874657595)(92, 0.2442959970653788)(93, 0.24427074113317127)(94, 0.24424600442825006)(95, 0.24422177110135224)(96, 0.24419802594178452)(97, 0.24417475434558444)(98, 0.24415194228558007)(99, 0.24412957628316623)};
    \addlegendentry{{$\beta=0.6$}}

\addplot[
    color=blue,smooth
    ]
    coordinates {
    (2, 0.2643996874222162)(3, 0.24894435717860763)(4, 0.23980150503669073)(5, 0.2337503697994383)(6, 0.22944696191200203)(7, 0.22622874271554266)(8, 0.22373082222364105)(9, 0.2217355416010075)(10, 0.22010496789678963)(11, 0.21874743270897468)(12, 0.21759962677207234)(13, 0.21661641630376344)(14, 0.21576475513618198)(15, 0.21501989090863166)(16, 0.21436291610392888)(17, 0.21377913864091352)(18, 0.21325696888005574)(19, 0.21278714163986923)(20, 0.21236216117975415)(21, 0.21197589798532873)(22, 0.21162329102212918)(23, 0.21130012461305792)(24, 0.2110028589920333)(25, 0.21072850004808386)(26, 0.21047449807652485)(27, 0.2102386682699648)(28, 0.21001912769096032)(29, 0.2098142448731814)(30, 0.2096225991942342)(31, 0.20944294787883067)(32, 0.20927419901101252)(33, 0.20911538931628176)(34, 0.20896566575815434)(35, 0.20882427020634858)(36, 0.20869052659465376)(37, 0.20856383010919732)(38, 0.20844363804220772)(39, 0.2083294620194277)(40, 0.20822086136635076)(41, 0.20811743742325706)(42, 0.2080188286543988)(43, 0.20792470642485816)(44, 0.2078347713410953)(45, 0.20774875006931165)(46, 0.20766639256040775)(47, 0.2075874696221468)(48, 0.20751177078891347)(49, 0.20743910244732722)(50, 0.20736928618257994)(51, 0.20730215731576296)(52, 0.20723756360690776)(53, 0.2071753641022711)(54, 0.2071154281074744)(55, 0.20705763427074947)(56, 0.20700186976277102)(57, 0.20694802954139313)(58, 0.20689601569121624)(59, 0.20684573682925922)(60, 0.206797107569127)(61, 0.20675004803705405)(62, 0.20670448343410053)(63, 0.20666034363941405)(64, 0.2066175628501115)(65, 0.2065760792539667)(66, 0.20653583473138515)(67, 0.20649677458373672)(68, 0.20645884728526737)(69, 0.20642200425635893)(70, 0.20638619965592164)(71, 0.20635139019106802)(72, 0.20631753494253652)(73, 0.20628459520415815)(74, 0.2062525343352673)(75, 0.20622131762472906)(76, 0.2061909121655769)(77, 0.20616128673927941)(78, 0.20613241170884222)(79, 0.20610425891989492)(80, 0.20607680160914377)(81, 0.2060500143194643)(82, 0.20602387282121715)(83, 0.2059983540391238)(84, 0.20597343598434564)(85, 0.20594909769126912)(86, 0.20592531915867013)(87, 0.20590208129492682)(88, 0.2058793658668438)(89, 0.20585715545198652)(90, 0.20583543339407456)(91, 0.20581418376135707)(92, 0.2057933913075949)(93, 0.20577304143565694)(94, 0.20575312016326086)(95, 0.2057336140909714)(96, 0.2057145103721726)(97, 0.20569579668479449)(98, 0.20567746120484154)(99, 0.20565949258147204)
    };
    \addlegendentry{{$\beta=0.8$}}

\addplot[
    color=green,smooth
    ]
    coordinates {
    (2, 0.24640234871610128)(3, 0.22699941778229937)(4, 0.21661474691570604)(5, 0.2101367190473883)(6, 0.2057076658588735)(7, 0.20248753940308178)(8, 0.2000405584438829)(9, 0.19811801442361732)(10, 0.19656757848306)(11, 0.19529072441072393)(12, 0.19422089620586414)(13, 0.19331151115894285)(14, 0.19252898904852156)(15, 0.19184851073477094)(16, 0.19125133554769963)(17, 0.19072304759708897)(18, 0.19025237616692559)(19, 0.18983038226400423)(20, 0.1894498852337)(21, 0.18910505065458472)(22, 0.18879108895391422)(23, 0.18850403152390888)(24, 0.18824056204228828)(25, 0.18799788774244425)(26, 0.18777364001342572)(27, 0.18756579681837704)(28, 0.18737262154101714)(29, 0.18719261433993445)(30, 0.1870244731243515)(31, 0.1868670620020046)(32, 0.1867193855815119)(33, 0.18658056789990185)(34, 0.18644983503233803)(35, 0.18632650065461076)(36, 0.18620995398956725)(37, 0.18609964969050452)(38, 0.18599509930788216)(39, 0.18589586405755815)(40, 0.18580154866472282)(41, 0.1857117961013507)(42, 0.18562628306944873)(43, 0.18554471610962264)(44, 0.18546682823624694)(45, 0.18539237601794556)(46, 0.1853211370360653)(47, 0.18525290766531582)(48, 0.1851875011298605)(49, 0.18512474579581506)(50, 0.18506448366726086)(51, 0.18500656905802987)(52, 0.1849508674157579)(53, 0.1848972542782114)(54, 0.18484561434485797)(55, 0.1847958406491016)(56, 0.18474783381866555)(57, 0.1847015014133651)(58, 0.18465675733101694)(59, 0.18461352127336988)(60, 0.18457171826519309)(61, 0.18453127822036997)(62, 0.18449213554976718)(63, 0.1844542288062519)(64, 0.18441750036279725)(65, 0.18438189612018063)(66, 0.1843473652410727)(67, 0.18431385990785326)(68, 0.18428133510168054)(69, 0.18424974840068198)(70, 0.18421905979533915)(71, 0.18418923151942518)(72, 0.18416022789495273)(73, 0.18413201518980565)(74, 0.18410456148685916)(75, 0.184077836563541)(76, 0.1840518117808379)(77, 0.18402645998091421)(78, 0.1840017553925713)(79, 0.18397767354390365)(80, 0.1839541911813697)(81, 0.18393128619496907)(82, 0.18390893754873938)(83, 0.183887125216379)(84, 0.18386583012135793)(85, 0.18384503408125913)(86, 0.18382471975602815)(87, 0.1838048705997425)(88, 0.18378547081567304)(89, 0.18376650531433436)(90, 0.18374795967433305)(91, 0.1837298201058471)(92, 0.18371207341637008)(93, 0.18369470697877738)(94, 0.1836777087012788)(95, 0.1836610669994213)(96, 0.18364477076972113)(97, 0.18362880936494236)(98, 0.18361317257097548)(99, 0.18359785058492567)
    };
    \addlegendentry{{$\beta=0.9$}}
    
\addplot[
    only marks,
    color = red,
    mark=square,
    error bars/.cd,
    y dir=both, y explicit,
    error bar style={color=red},
    ]
    coordinates {
        (2, 0.30580999999999997) +- (2, 0.04276410761374544)
        (4, 0.30302) +- (4, 0.023023240432224146)
        (6, 0.30434) +- (6, 0.030269694415371973)
        (8, 0.29436) +- (8, 0.02260310598125842)
        (10, 0.29086) +- (10, 0.0212788721505629)
        (12, 0.29596999999999996) +- (12, 0.02822789577705004)
        (14, 0.28112999999999994) +- (14, 0.021016614855870586)
        (16, 0.28985000000000005) +- (16, 0.027476762909775243)
        (18, 0.2837) +- (18, 0.02278047409515438)
        (20, 0.28097) +- (20, 0.01727611356758224)
        (22, 0.27343999999999996) +- (22, 0.01774081170634534)
        (24, 0.27423000000000003) +- (24, 0.014980724281555946)
        (26, 0.27213) +- (26, 0.017814828093473123)
        (28, 0.2726) +- (28, 0.018074844397670486)
        (30, 0.27249) +- (30, 0.01968742999987555)
        (32, 0.27329) +- (32, 0.019395383471331518)
        (34, 0.26517) +- (34, 0.015277633979121241)
        (36, 0.26298) +- (36, 0.014114517349169242)
        (38, 0.26797000000000004) +- (38, 0.014320408513726124)
        (40, 0.26403000000000004) +- (40, 0.015505937572426909)
        (42, 0.26653999999999994) +- (42, 0.015792352579650726)
        (44, 0.26485) +- (44, 0.014308127061219422)
        (46, 0.26411) +- (46, 0.014508028811661522)
        (48, 0.26441) +- (48, 0.011748910587794925)
        (50, 0.26682999999999996) +- (50, 0.012280313513913232)
    };
    
\addplot[
    only marks,
    color =blue,
    mark=square,
    error bars/.cd,
    y dir=both, y explicit,
    error bar style={color=blue},
    ]
    coordinates {
    (2, 0.28764) +- (2, 0.035517043795901704)
    (4, 0.26391) +- (4, 0.02089705481640891)
    (6, 0.26035) +- (6, 0.018464357557196498)
    (8, 0.24373) +- (8, 0.01657504449466124)
    (10, 0.2386) +- (10, 0.011033585092797366)
    (12, 0.23402000000000003) +- (12, 0.014939531451822724)
    (14, 0.22644000000000003) +- (14, 0.01021598747062663)
    (16, 0.22880999999999996) +- (16, 0.010762848136065095)
    (18, 0.22772000000000003) +- (18, 0.010029536380112503)
    (20, 0.22328999999999996) +- (20, 0.008626175282244158)
    (22, 0.22332999999999997) +- (22, 0.011588619417342183)
    (24, 0.22096) +- (24, 0.009022660361556356)
    (26, 0.22023999999999996) +- (26, 0.008792747011031304)
    (28, 0.21676) +- (28, 0.007979498731123413)
    (30, 0.22087) +- (30, 0.009738485508537786)
    (32, 0.21672) +- (32, 0.007641439654934162)
    (34, 0.21789999999999995) +- (34, 0.009641058033224376)
    (36, 0.21811) +- (36, 0.009289397181733605)
    (38, 0.21828999999999998) +- (38, 0.010569810783547656)
    (40, 0.21601999999999996) +- (40, 0.007799974358932232)
    (42, 0.21713000000000002) +- (42, 0.008427817036457314)
    (44, 0.21328) +- (44, 0.008533557288727844)
    (46, 0.21496999999999997) +- (46, 0.007489599455244578)
    (48, 0.2152) +- (48, 0.008500588214941374)
    };


\addplot[
    only marks,
    color =green,
    mark=square,
    error bars/.cd,
    y dir=both, y explicit,
    error bar style={color=green},
    ]
    coordinates {
        (2, 0.2648) +- (2, 0.021147813125711123)(4, 0.22388000000000002) +- (4, 0.015288740955356678)(6, 0.21617999999999996) +- (6, 0.015955801452763207)
        (8, 0.20745999999999998) +- (8, 0.010999927272486859)
        (10, 0.20325) +- (10, 0.01023877434071091)
        (12, 0.19841999999999996) +- (12, 0.00877858758571104)
        (14, 0.19554) +- (14, 0.008577202341089996)
        (16, 0.19558999999999999) +- (16, 0.009215036624995038)
        (18, 0.19297) +- (18, 0.00629556192885114)
        (20, 0.19433) +- (20, 0.009929556888401414)
        (22, 0.19258) +- (22, 0.005892673417049327)
        (24, 0.19116) +- (24, 0.008909792365706382)
        (26, 0.19114) +- (26, 0.008508254815178001)
        (28, 0.19012) +- (28, 0.006861166081651104)
        (30, 0.18858999999999998) +- (30, 0.006260742767435823)
        (32, 0.19182999999999997) +- (32, 0.007776638091103377)
        (34, 0.18746000000000002) +- (34, 0.006213565803948634)
        (36, 0.18739999999999998) +- (36, 0.006549045732013148)
        (38, 0.18802000000000002) +- (38, 0.007507969099563444)
        (40, 0.1889) +- (40, 0.005693856338194692)
        (42, 0.18692) +- (42, 0.006234067692927281)
        (44, 0.18720000000000003) +- (44, 0.007125307010929421)
        (46, 0.18593000000000004) +- (46, 0.006551343373690621)
        (48, 0.18761999999999998) +- (48, 0.007667828897412854)
        (50, 0.18627999999999997) +- (50, 0.005824225270368577)
        (52, 0.18634) +- (52, 0.006841374130976879)
    };

    
    
\end{axis}
\end{tikzpicture}
    \vspace*{-.2cm}
\caption{Theoretical (solid lines) versus empirical (markers) error rates when adding source tasks with relatedness $\beta$ to target task. {\bf Marked improvement on first added tasks before asymptotic saturation.} }
\label{fig:added}
\end{figure}

We next study the increase of the number $k-1$ of source ($2$-class) tasks, first on synthetic Gaussian and then on real data. For synthetic data, in Figure~\ref{fig:added}, $x_{tl}^{(j)}\sim\mathcal{N}((-1)^j\mu_t,I_p)$, $\mu_t=\beta\mu+\sqrt{1-\beta^2}\mu^{\perp(t)}$ with $\beta\in[0,1]$ fixed, $\mu=e_1^{[p]}$ and $\mu^{\perp(t)}$ random of unit norm with $[\mu^{\perp(t)}]_1=0$, for $t\in\{1,\ldots,k\}$. Tasks are successively added to help classify the target task (task 2) with $n_{t1}=n_{t2}=50$ for $t\ne2$ and $n_{21}=n_{22}=20$, and $p=100$. Not surprisingly, larger values of $\beta$ induce higher performance levels with an ultimate saturation as $k\to\infty$. The slight mismatch between theory (from \cite{MTLSPCA}) and practice decreases as $n,p\to\infty$, revealing a rather unexpected phenomenon: that additional degrees of freedom further help the algorithm to ``discover'' and exploit the affinity between tasks.  

\medskip

Real data arise from Amazon Review (textual user reviews, positive or negative, on books, DVDs, electronics, and kitchen items) and Office31+Caltech (two out of ten classes of images from different modalities: Amazon, Caltech-256, high-resolution DSLR, or low-resolution webcam images) \cite{blitzer-etal-2007-biographies,officeCal}.  Figure~\ref{fig:amazon} depicts the classification performance focusing on the ``kitchen'' task (task 1) by successively adding the three other datasets as sources in the order given in the figure, for $n_{11}=n_{12}=n_{31}=n_{32}=200$, $n_{21}=n_{41}=50$, $n_{22}=n_{42}=90$, $p=400$.
\begin{figure}[h!]
\centering
\begin{tikzpicture}
\begin{axis}[
    width=\linewidth,height=.45\linewidth,font=\footnotesize,
    legend style={anchor=north east,at={(.98,.98)},font=\footnotesize,fill opacity=0.7, draw opacity=1},    
    xtick = {2,4,6},
    xticklabels = {Books,DVDs,Electronics},
    xlabel={Added Tasks},
    ylabel={Error rate},
    ymajorgrids=true,
    xmajorgrids=true,
    grid style=dashed,
]

\addplot[
    color=blue,
    mark=square,
    ]
    coordinates {
    (2, 0.20060030015007502)(4, 0.19509754877438723)(6, 0.18509254627313654)
    };
    \legend{Distributed MTL-SPCA}

\addplot[
    color=red,
    mark=triangle,
    ]
    coordinates {
    (2, 0.216608304152076)(4, 0.216608304152076)(6, 0.216608304152076)
    };
    \addlegendentry{Single-task SPCA}

\addplot[
    color=black,
    mark=x,
    ]
    coordinates {
    (2, 0.2041020510255127)(4, 0.20060030015007502)(6, 0.19309654827413703)
    };
    \addlegendentry{N-SPCA}
    
\end{axis}
\end{tikzpicture}
    \vspace*{-.5cm}
\caption{Classification error for the Amazon Review dataset with ``Kitchen'' as target task under successive addition of sources. MTL-SPCA (thus with optimized labels $\tilde{y}$) versus single-task SPCA (no source data) and the ``naive'' (N-PCA) algorithm with labels $y_i\in\{\pm1\}$. {\bf Transfer learning significantly helps with a clear improvement induced by optimal labels.} }
\label{fig:amazon}
\end{figure}
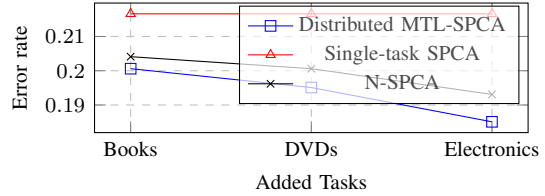
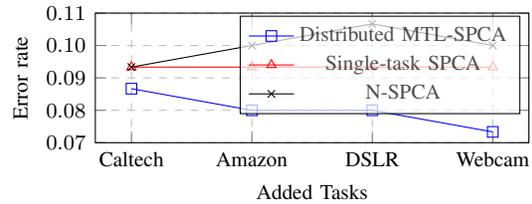
\begin{figure}[h!]
\centering
\begin{tikzpicture}
\begin{axis}[
    xtick = {2,4,6, 8},
    width=\linewidth,height=.45\linewidth,font=\footnotesize,
    legend style={anchor=north east,at={(.98,.98)},font=\footnotesize,fill opacity=0.7, draw opacity=1},      
    xticklabels = {Caltech,Amazon, DSLR, Webcam},yticklabel style={%
        /pgf/number format/fixed,
        /pgf/number format/precision=2,
        /pgf/number format/fixed zerofill
    },
    xlabel={Added Tasks},
    ylabel={Error rate},
    ymajorgrids=true,
    xmajorgrids=true,
    grid style=dashed,
]

\addplot[
    color=blue,
    mark=square,
    ]
    coordinates {
    (2, 0.086666)(4, 0.08)(6, 0.08)(8, 0.07333)
    };
    \addlegendentry{Distributed MTL-SPCA}
   
\addplot[
    color=red,
    mark=triangle,
    ]
    coordinates {
    (2, 0.0933333)(4, 0.0933333)(6, 0.0933333)(8,0.0933333)
    };
    \addlegendentry{Single-task SPCA} 

\addplot[
    color=black,
    mark=x,
    ]
    coordinates {
    (2, 0.0933333)(4, 0.1)(6, 0.10666)(8,0.1)
    };
    \addlegendentry{N-SPCA} 
    
\end{axis}
\end{tikzpicture}
    \vspace*{-.3cm}
\caption{Classification error for the Office+Caltech dataset, targetting Caltech images of classes ``Backpack'' and ``Touring-Bike''; same algorithms as in Figure~\ref{fig:amazon}. {\bf Clear advantage of MTL-SPCA and a strong manifestation of the ``negative transfer'' effect for N-SPCA.} }
\label{fig:office31}
\end{figure}
SURF features ($p=800$) are considered as input data for Office+Caltech images. In Figure~\ref{fig:office31}, features from Amazon (task 2), DSLR (task 3), then webcam images (task 4) are added to classify the ``Backpack'' and ``Touring-Bike'' classes of Caltech (task 1), for $n_{11}=50$, $n_{12}=60$, $n_{21}=90$, $n_{22}=80$, $n_{31}=n_{32}=12$, $n_{41}=n_{42}=20$. 

Both figures demonstrate the overall superiority of MTL-SPCA. Most importantly, while N-SPCA may severely suffer from negative transfer (performing worse than with no extra task), MTL-SPCA both ignores the negative transfer problem but also improves where N-SPCA decays.

\section{Conclusion and discussion}

MTL-SPCA is one example of a simple answer to the difficult multi-task learning problem, yet a solution which is (i) optimal in the (rather large) Gaussian isotropic case, (ii) easy to implement in a distributed manner, and (iii) cost-efficient and environmentally friendly when compared to modern machine learning mechanisms. What it only takes here to make MTL-SPCA so powerful is a statistical analysis and improvement by random matrix theory \cite{MTLSPCA}, a comparison to the information-theoretic optima, and a thorough inspection of the incompressible sufficient statistics at play in view of a distributed implementation.

We believe that a more systematic ``sober'' approach to ML problems, based on modern mathematical techniques, is prone to significantly reduce the environmental footprint of ML algorithms, which is becoming an absolute necessity.


\newpage

\bibliographystyle{plain}
\bibliography{biblio}

\end{document}